\pdfoutput=1
\documentclass[letterpaper, 10 pt, journal, twoside]{IEEEtran}  

\IEEEoverridecommandlockouts                              

\usepackage{amsmath}
\usepackage[linesnumbered,algoruled,boxed,lined]{algorithm2e}
\usepackage[noend]{algpseudocode}
\usepackage{amssymb}
\usepackage{amsmath}
\usepackage{hyperref}
\usepackage{graphicx}
\usepackage{caption}
\usepackage{subcaption}
\graphicspath{ {./images/} }
\usepackage{color}
\usepackage[table]{xcolor}
\usepackage{url}
\usepackage{leftidx}
\usepackage[noadjust]{cite}
\usepackage{filecontents}
\usepackage{booktabs}
\usepackage{soul}
\usepackage{array}
\usepackage{ulem}
\usepackage{tikz}
\usepackage{boldline}

\usepackage{graphics} 
\usepackage{booktabs}
\usepackage{multirow}
\usepackage{threeparttable}
\usepackage{overpic}
\usepackage{makecell}

\usetikzlibrary{positioning}
\usepackage[top=0.75in, bottom=0.7in, left=0.7in, right=0.7in]{geometry}
\newcommand{\comment}[1]{{\color{black}{#1}}}

\usepackage{microtype}

\title{Learn to Grasp via Intention Discovery and its Application to Challenging Clutter}

\author{Chao Zhao, Chunli Jiang, Junhao Cai, Hongyu Yu, Michael Yu Wang, and Qifeng Chen

\thanks{Manuscript received: September, 5, 2022; Revised November, 15, 2022; Accepted December, 4, 2022. }

\thanks{This paper was recommended for publication by Editor Hong Liu upon evaluation of the Associate Editor and Reviewers’ comments.
This work was supported by the Project of Hetao Shenzhen-Hong Kong Science and Technology Innovation Cooperation Zone, HZQB-KCZYB-2020083. }

\thanks{C. Zhao, C. Jiang, J. Cai, H. Yu, and Q. Chen are with The Hong Kong University of Science and Technology, Clear Water Bay, Hong Kong  {\tt\footnotesize \{czhaobb, cjiangab, jcaiaq\}@connect.ust.hk} and {\tt\footnotesize \{hongyuyu, cqf\}@ust.hk}. J. Cai, H. Yu, and M. Wang are also with HKUST Shenzhen-Hong Kong Collaborative Innovation Research Institute, Futian, Shenzhen. M. Wang is with Monash University {\tt\footnotesize michael.y.wang@monash.edu}}

\thanks{Digital Object Identifier (DOI): see top of this page.} 
}

\begin{document}

\maketitle

\begin{abstract}

Humans excel in grasping objects through diverse and robust policies, many of which are so probabilistically rare that exploration-based learning methods hardly observe and learn. Inspired by the human learning process, we propose a method to extract and exploit latent intents from demonstrations, and then learn diverse and robust grasping policies through self-exploration. The resulting policy can grasp challenging objects in various environments with an off-the-shelf parallel gripper. The key component is a learned intention estimator, which maps gripper pose and visual sensory to a set of sub-intents covering important phases of the grasping movement. Sub-intents can be used to build an intrinsic reward to guide policy learning. The learned policy demonstrates remarkable zero-shot generalization from simulation to the real world while retaining its robustness against states that have never been encountered during training, novel objects such as protractors and user manuals, and environments such as the cluttered conveyor. 

\end{abstract}

\section{Introduction}
\begin{figure}[!t]
    \centering
    \begin{overpic}[width=\linewidth]{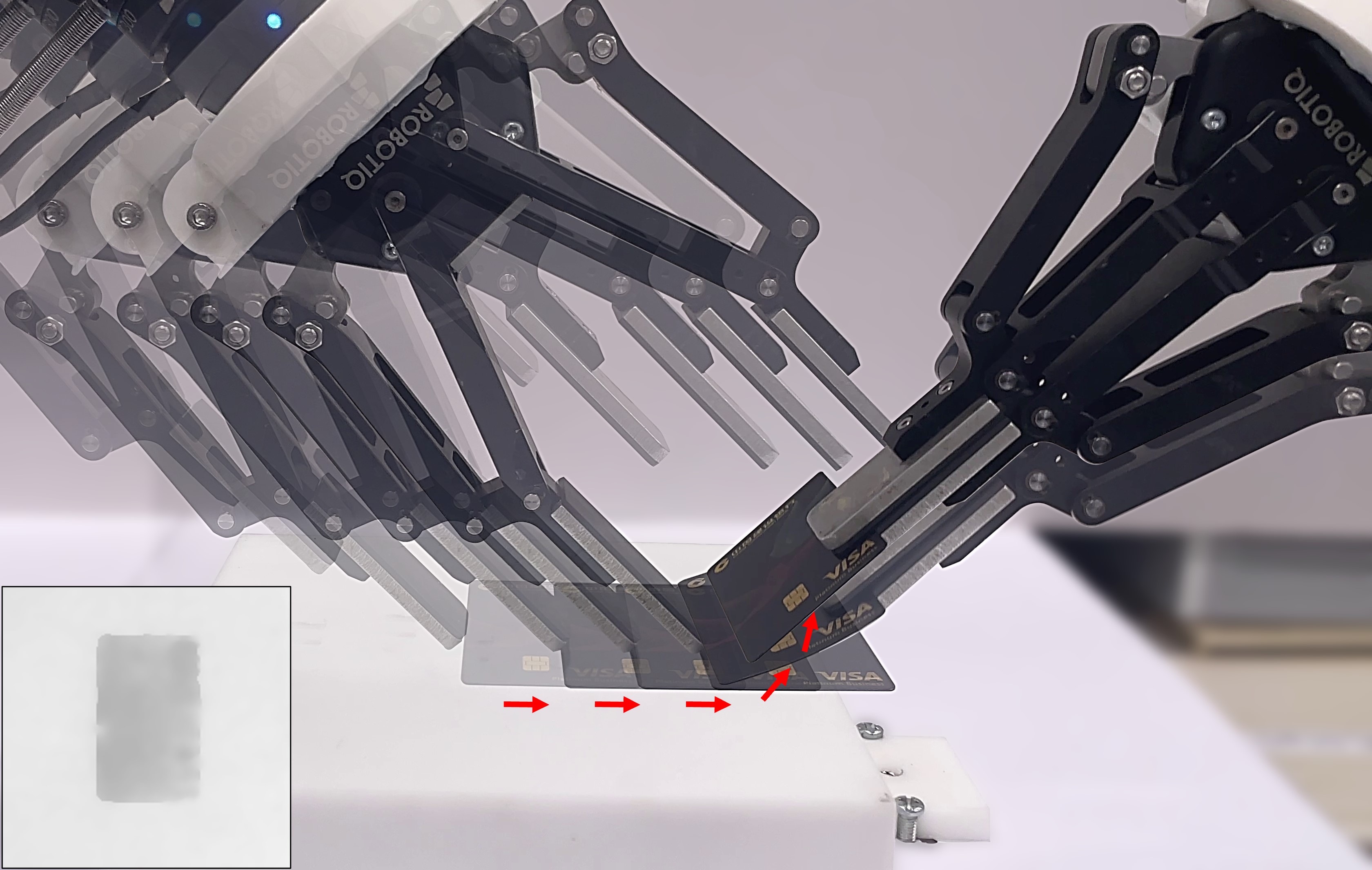}
    \end{overpic}
    \caption{A parallel gripper with the learned policy picking objects using vision for sensing. The time-lapse image shows the actions of the gripper as it interacts with a credit card to pick it up. The red arrow shows the card's motion. The depth image on the bottom-left shows the visual observation. 
    }
    \label{fig:fg1}
\vspace{-0.5cm}
\end{figure}

Grasping is a fundamental maneuver in many tasks, and grasping a particular object may require a dedicated policy. For example, consider a common grasping scenario where the robot needs to grasp the credit card with a parallel gripper, as shown in Fig. \ref{fig:fg1}. Grasping a credit card object is challenging because the card is so thin that a successful grasp policy may require the gripper to interact with the object and utilize external surfaces to aid manipulation. Although developing flexible and robust policies for grasping diverse objects is a breeze for humans, the current state of the art in robotics is still far from such a capability. 

Recent studies have focused on autonomous grasping policy discovery. This area is dominantly driven by model-free reinforcement learning (RL), which obtains grasping policies by self-exploration \cite{pmlr-v87-kalashnikov18a, Bodnar-RSS-20}. However, an important issue with exploration-based methods is that some grasping policies are probabilistically rare, which results in the discovered grasping policies having a similar pattern (i.e., approaching the object and closing the fingers). In this regard, imitation learning offers a way to learn robot skills by mimicking the expert behaviors in demonstrations \cite{dyrstad2018teaching,johns2021coarse}. However, existing methods only attempt to match the expert action sequences \cite{florence2022implicit}, ignoring the understanding of high-level goal planning in the demonstration. As a result, the learned policy cannot be transferred to scenes absent from the demos, limiting the generalization ability. 

Therefore, exploiting dexterous grasping strategies from human demos while retaining the ability to explore and adapt to novel scenarios autonomously remains an open problem. This motivates us to propose a method inspired by the human learning process to address this challenge.

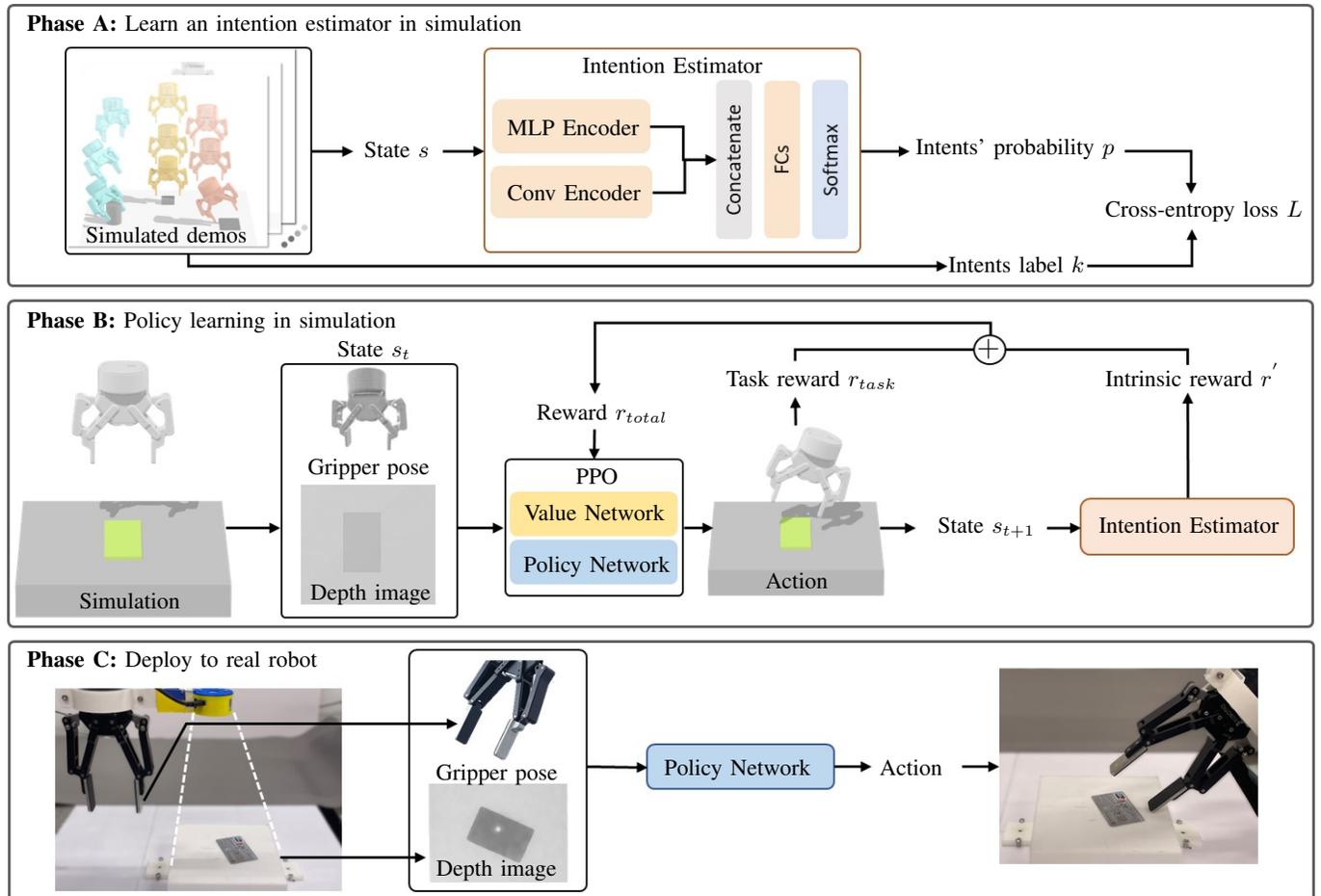
\begin{figure*}[!t]
\vspace{0.25cm}
    \centering
    \begin{overpic}[width=\linewidth]{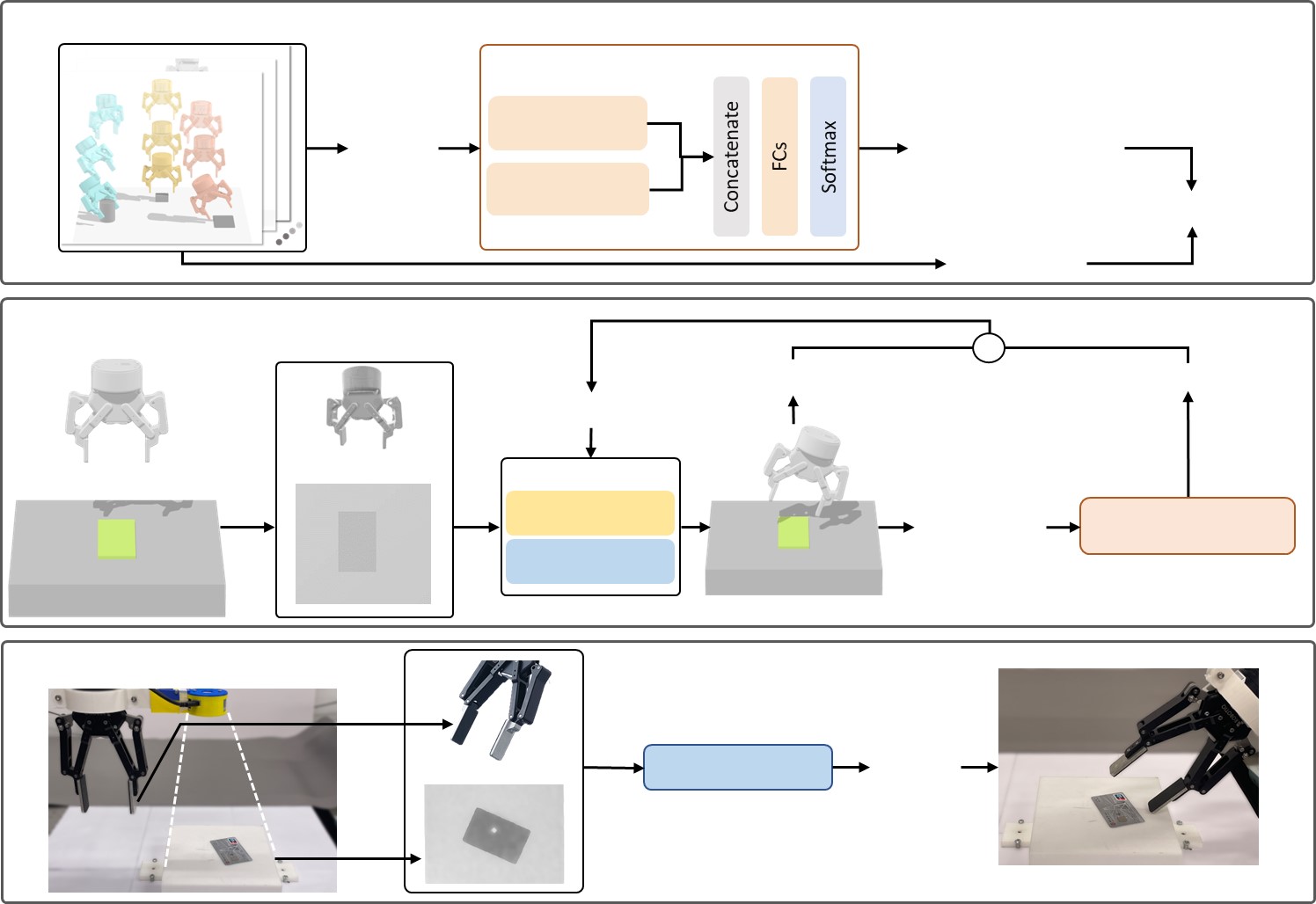}
    \put(1.5,66.75) {\small {\textbf{Phase A:} Learn an intention estimator in simulation }}
    \put(6,50.5) {\small {Simulated demos}}
    \put(27.3,57) {\small {State $s$}}
    \put(38.25,58.75) {\small {MLP Encoder}}
    \put(38.25,53.75) {\small {Conv Encoder}}
    \put(44,63.5) {\small {Intention Estimator}}
    
    \put(69.5,57.25) {\small {Intents' probability $p$}}
    \put(72,48.25) {\small {Intents label $k$}}
    \put(84,52.5) {\small {Cross-entropy loss $L$}}
    
    \put(1.5,44) {\small {\textbf{Phase B:} Policy learning in simulation}}
    \put(5.5,22.5) {\small {Simulation}}
    \put(25.25,41.75) {\small {State $s_t$}}
    \put(23,32.75) {\small {Gripper pose}}
    \put(23.2,23.15) {\small Depth image}
    
    \put(40.5,36.95) {\small {Reward $r_{total}$}}
    \put(43.5,32.05) {\small {PPO}}
    \put(39.55,29.25) {\small Value Network}
    \put(39.5,25.25) {\small {Policy Network}}
    
    \put(58,24) {\small {Action}}
    \put(71.2,28.25) {\small {State $s_{t+1}$}}
    
    \put(83.5,28.25) {\small Intention Estimator}   
    \put(84,39.5) {\small Intrinsic reward $r^{'}$}
    \put(55,39.5) {\small Task reward $r_{task}$}
    \put(73.5,41.75) { \large \textbf{$+$}}

    \put(1.5,18) {\small {\textbf{Phase C:} Deploy to real robot}}
    
    \put(32.825,2) {\small Depth image}
    \put(32.8,9.35) {\small {Gripper pose}}
    
    \put(50.25,9.75) {\small {Policy Network}}
    \put(66.75,9.75) {\small {Action}}

    \end{overpic}
    \caption{\textbf{System Overview.} \textbf{A:} We generate a set of simulated grasps to learn an intention estimator. The state $s$ in a grasp includes the depth image and gripper pose. They are processed separately using a Conv encoder for the former and an MLP encoder for the latter. Then, the concatenation of two vectors is fed through the subsequent FC layers to predict probabilities. \textbf{B:}  We train our policy with PPO. The RL agent receives the observed state $s_t$ and predicts the action at time step $t$. The robot executes and switches to the next state $t+1$. The intention estimator discerns the intent of the given state $t+1$, and the RL agent then receives a task reward from the simulation and an intrinsic reward from the intention estimator. \textbf{C:} To transfer to the real world, the policy network alone is used to control the robot. The wrist-mounted camera provides the depth image, and the gripper pose is from the robot's proprioception.
    }
    \label{fig:fg3}
\vspace{-0.5cm}
\end{figure*}

Evidence from neuroscience suggests that when humans learn a skill or children learn from others, they selectively focus on the underlying intents of an actor's behavior rather than learning atomic actions \cite{meyer2020intention}. Then, learning is facilitated by following the intents and self-practice. Inspired by this intuitive introspection, we propose a framework to mimic this process to learn grasping, as shown in Fig. \ref{fig:fg3}. At its core, policy learning is based on a principled solution to incorporate the intrinsic reward from intents into RL training. The key component is an intention estimator that predicts probability distributions of a set of intents.
The intents are the temporal abstraction of the important phases in the grasping trajectories (e.g., go to a position, rotate, close gripper) compared with detailed movements. The RL agent leverages the foresight afforded by the intention estimator to guide policy learning. Meanwhile, the agent is able to learn policies purely by self-exploration when the intention estimator meets novel scenes.
Thus, the proposed approach combines the best of both worlds: the diversity of policies provided by demonstration and the adaptability and generalizability brought by self-exploration.

The primary contribution of this paper is the proposed method for learning dexterous grasping policies that have the ability to:
1) grasp objects in broad categories, such as credit cards, Go stones, and soda cans, with only an off-the-shelf parallel gripper;
2) grasp in scenes that are never encountered in demonstrations, such as a cluttered table;
3) learn grasping policies that are unable to be obtained purely by self-exploration;
and 4) only use easy, readily available sensors such as the depth camera.
While some of these features have been individually demonstrated, we are unaware of published studies that tackle all four.

\section{Related Work}

Grasping is a fundamental problem in robotics; it enables further in-hand manipulation and interaction with the environment. Conventional analytic methods model physical processes between the object and the gripper and use model-based planning to output grasping policies \cite{bicchi2000robotic}. However, the complexity of physical analysis typically assumes known physical properties to make analysis or planning tractable, which are hard to obtain in practice. \comment{Meanwhile, objects are often occluded in cluttered scenes, which makes analyzing feasible grasps challenging. Some works have explored using pre-grasp manipulation, such as sliding \cite{pre1}, to create graspable poses of objects. For example, \cite{pre2} grasp an object by pushing the object against a support surface and lift the object by pivoting. However, such methods need prior knowledge of the environment and objects' physical properties.} 

Learning-based methods have recently emerged as alternatives to robotic grasping \cite{newbury2022deep, r2}, as they can detect grasps from visual features rather than explicitly using prior knowledge of objects. For example, some grasp synthesis methods \cite{ten2017grasp, cai2022real} use neural networks to accept visual observation as input and output pose estimates of feasible grasps. On the other hand, \cite{zeng2018robotic} defines each pixel as a top-down grasp primitive rather than predicting a grasp pose and evaluates each grasp quality through a fully-convolutional neural network. \cite{learntodig2022} extends this method with an adjustable finger and a model-based primitive to produce an effective grasping system. While predefined primitives can improve data efficiency, they also limit the diversity of policies. Another line of work uses model-free RL algorithms to acquire the grasping policy autonomously through self-exploration \cite{ibarz2021train, cao2022reinforcement,pmlr-v87-kalashnikov18a}. 
However, the grasping policy is often hard to explore, particularly as the degrees of freedom increase \cite{morel2022automatic}. 
\comment{Some studies \cite{cluster1, cluster2} introduce clustering-based intrinsic rewards to accelerate RL learning but cannot obtain policies beyond self-exploration capabilities.}

For learning dexterous policies, imitation learning is a common approach \cite{song2020grasping,dyrstad2018teaching, hamaya2020learning}. The well-known imitation learning method includes behavior cloning, which realizes a mapping between robot states and actions from human demonstration. However, the application of imitation learning to grasping has largely been confined to the quality of expert demonstrations, and collecting demonstration data is often expensive and time-consuming \cite{song2020grasping}. Moreover, the common issue with these methods is that they are hard to generalize to unseen objects or environments that are not included in demonstrations due to distributional shifts and compounding errors \cite{florence2022implicit},\cite{r3}. Other works focus on designing end-effectors to grasp challenging objects instead of focusing on the grasping policy. The end effector can be designed by humans or discovered through learning algorithms \cite{kodnongbua2022computational}. However, end effectors with complicated designs often only apply to specific object types, reducing the robot's versatility and increasing the system's complexity.

Compared with the abovementioned studies, our presented approach substantially improves the diversity of graspable objects and the grasping policies. Rather than imitating atomic actions, our method extracts the latent intents from demos and utilizes them in policy learning, incorporated with self-exploration. The entire learning is completed in simulation without expensive demo collection in field conditions and is consistently effective in zero-shot transfer to the real world.

\section{Method}\label{sec:method}

In this section, we describe the proposed method for learning the grasping policy. Our method consists of three phases,  as illustrated in Fig. \ref{fig:fg3}. 

First, an intent estimator is trained with simulated grasps to learn a mapping between the state in a grasp demonstration and intents (see Sec. \ref{sec:learn_intention}). Grasps are generated in the simulation using the three provided grasp types (see Fig. \ref{fig:fg2}). The intention estimator captures the environment and robot information using a network and outputs a set of probabilities representing the distance between the given state and intents. 

In the second phase, the grasping policy is trained with RL (see Sec. \ref{sec:policy_rl}). During training, we exploit two kinds of rewards: a task reward and an intrinsic reward. The task reward is sparse and given when the robot successfully grasps. The intrinsic reward is from the intention estimator and guides the RL agent when the agent approaches an intent. Chronologically, the latter appeared intention is both achievable and closer to the solution than the former. Therefore, providing positive rewards can facilitate robot learning after each intent is fulfilled. Such construction compensates for the inability to discover interesting policies with random exploration.

Last, we transfer and deploy the learned policy to the physical robot. Our training in simulation only uses rigid objects with simple geometry, such as the cube and cylinder (see Sec. \ref{sec:DC+AB_METHOD}). Yet when deployed on a real robot, the robot successfully handles broad object categories (protractors, Go stones, etc.) and environments (cluttered table and conveyor) with only an off-the-shelf parallel gripper. 

\subsection{Learning an Intention Estimator}\label{sec:learn_intention}

\begin{figure}[!tbp]
\vspace{0.25cm}
    \centering
    \begin{overpic}[width=\linewidth]{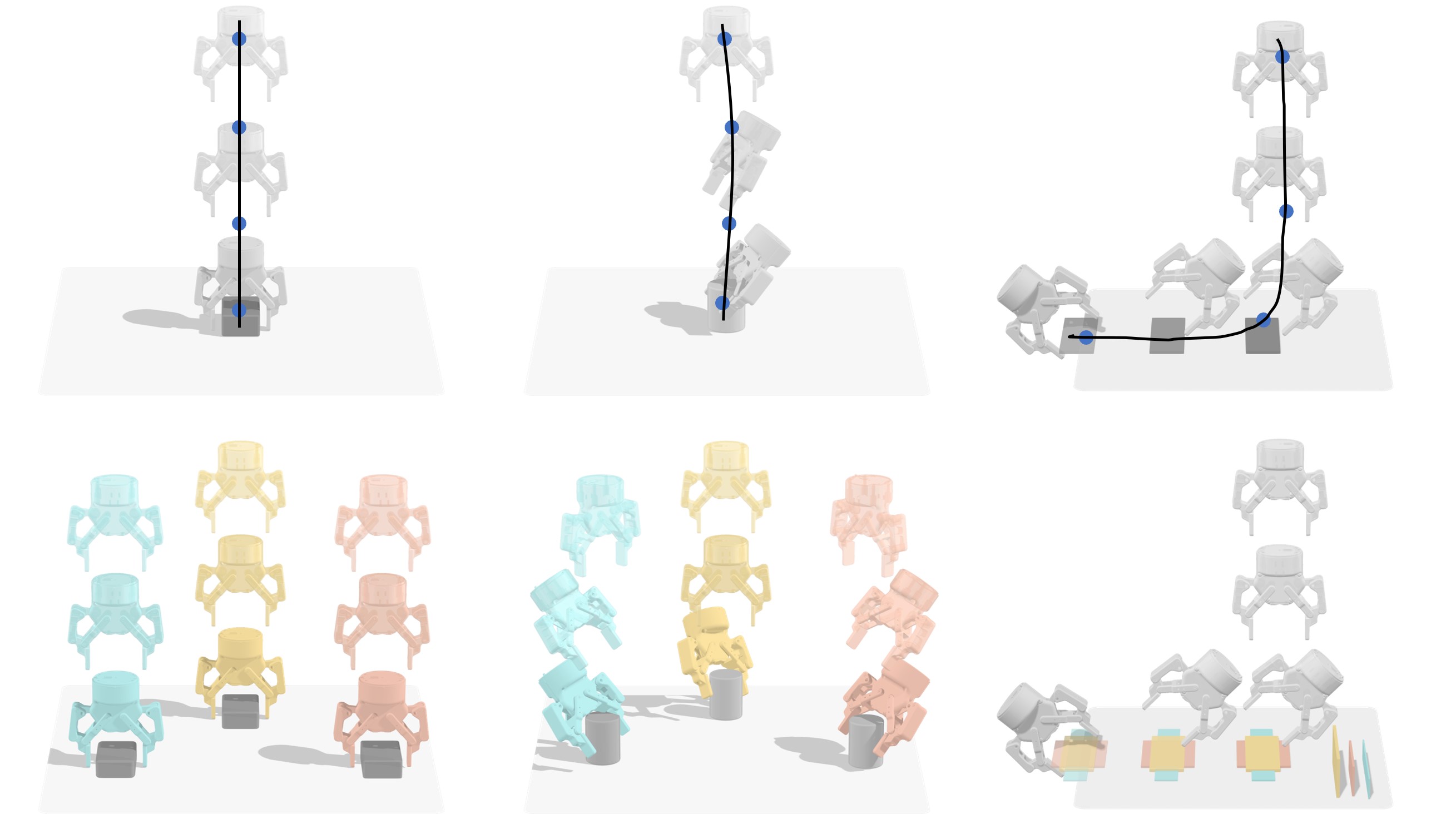}

    \put(48,28){\scriptsize  {(a)}}
    \put(48,-1){\scriptsize  {(b)}}
    
    \end{overpic}
    \caption{\comment{Data collection for learning the intention estimator.} (a) Three demonstrated grasps;  (b) \comment{Examples of grasps augmented based on the three demos in (a).} Left: changes in object positions. Middle: changes in the object orientations. Right: changes in object sizes.}
    \label{fig:fg2}
\vspace{-0.5cm}
\end{figure}

In the first phase of learning, we aim to learn an intention partitioning strategy with a neural network, as shown in Fig. \ref{fig:fg2}A. The input is a given state from simulated grasp demos, and the output is a family of probability distributions indicating how likely the current state is to be divided into each intent.

\textbf{Intent Segmentation and Data Collection:} Considering a grasp demonstration $S=(s_{1}, s_{2}, \ldots, s_l)$ represented by a sequence of states $s = (I, h)$, where $I$ is the camera observation of the environment, and $h$ is the gripper pose. The state $s \in S$ in a grasp demo can be naturally segmented into $n$ intents, denoted as $K = (k_0, k_1, \ldots, k_n)$, according to the timing order and similarity. The index of $k$ indicates the timing order of intents, and an intent $k_{t+1}$ can only be reached after completing former intent $k_{t}$. The $h$ consists of $ (x,y,z,\alpha,\beta,\gamma, \psi)$, where $(x,y,z, \alpha,\beta,\gamma) \in SE(3)$ is the 6D gripper pose, and $\psi$ is one hot vector representing the opening and closing of the gripper. 

We now give a formal definition of the $k$-intent segmentation problem. If $k = l$, each gripper pose corresponds to a segment. Otherwise, despite the fact that humans can manually label segments, the following segmentation algorithm can be used to reduce labor costs. Let $T=(T_0, T_1,\ldots)$ denote the set of all possible ways of segmentation for a sequence $S$. The sequence $S$ of length $l$ contains $n$ non-overlapping contiguous sub-sequences, denoted as $T_i = (\tau_1, \tau_2, . . . , \tau_n)$. Each state $s$ in segment $\tau_i$ belongs to the intent $k_i$. We denote the dissimilarity in a segment $\tau_i$ as $e_i$, then the error of segmentation $T_i$ is calculated as $E_{p}=\sum_{i=0}^{n} e_{i}$. Thus, we define the optimal segmentation as to find the minimize $E_{p}$ in $T$:
\begin{equation}
T_{\mathrm{opt}}(S, n)=\arg \min _{T_i \in T} E_{p}(S, T_i).
\end{equation}
The $T_{\mathrm{opt}}(S, n)$ can be found by the dynamic-programming (DP) algorithm \cite{dp}, and the main recurrence of the DP is
\begin{equation}
\begin{aligned}
E_p\left(T_{\mathrm{opt}}\right.&(S[1 \ldots l], n))= \left\{E_p\left(T_{\mathrm{opt}}(S[1 \ldots j], n-1)\right)\right.\\
&\left.+E_p\left(T_{\mathrm{opt}}(S[j+1, \ldots, l], 1)\right)\right\},
\end{aligned}
\end{equation}
where $S[1, \dots, j]$ denotes the sub-sequence of $S$ that contains states in positions from $1$ to $j$. \comment{The function of the recurrence is to divide the segmentation problem into subproblems and combine their solutions to form the final segmentation.} The dissimilarity $e_i$ in a segment is the sum of the  dissimilarities $\Lambda$ between states, calculated by the following formula:
\begin{equation}
e_{i}=\sum_{s_{v}, s_{w} \in\left(\begin{array}{c}
s \in \tau_i \\
2
\end{array}\right)} |\Lambda_{s_v, s_w}|,
\end{equation}
\begin{equation}
\begin{aligned}
|\Lambda_{s_v, s_w}|
&=(|x_{v}-x_{w}|+|y_{v}-y_{w}| +|z_{v}-z_{w}|) \\
&+(|\alpha_{v}-\alpha_{w}|+|\beta_{v}-\beta_{w}| +|\gamma_{v}-\gamma_{w}|) \\
&+\mu(|\psi_{v}-\psi_{w}|),
\end{aligned}
\end{equation}
where $\lambda$ and $\mu$ are the hyper-parameters to adjust the influence of the gripper orientation and finger condition (i.e., open/close) change. \comment{The distance of the orientation is the relative difference of Euler angle changes and is normalized.} After segmentation, each state $s$ in a demo is assigned to an intent $k_i=(1,2,\ldots, n)$ as supervision signals.

To learn an intention estimator, a set of grasp demonstrations needs to be collected and segmented using the above algorithm. We generate grasps in simulation by augmenting three human-encoded grasps (see Fig. \ref{fig:fg3}(a)), using invariant and equivariant principles. Consider an encoded grasp $S$ for an object $o$ with a pose $o_p$. A new grasp $S^{\prime}$ can be augmented by the following procedures. \comment{First, we apply a set of transformations to the object pose $o_p$, including changing object positions and orientations. Then the new grasp $S^{\prime}$ is transferred from $S$ via homogeneous transformation by calculating the $SE(3)$ matrix between $o_p$ and $o^{\prime}_p$, as shown in Fig. \ref{fig:fg3}(b).} We also randomize the aspect ratio of objects at each new grasp generation. Although such augmentation of grasps leads to some imperfect grasps, these imperfect demos do not affect policy learning. We further analyze the influence of imperfect grasp demonstrations in Sec. \ref{sec: simulation results} and Fig. \ref{fig:fg4}(c).

\textbf{Intention Estimator Learning:} The goal of the intent estimator is to map the similarity between the given state and each intent. Operationally, we form it as a classification problem and use a neural network $p = f(s)$ to learn this mapping, where the given state $s$ contains a depth image $I$ of the environment and gripper pose $h$. The network processes grasp pose and visual observation in separate channels, and the output features are combined to feed into a feed-forward pipeline to calculate probabilities that the given state belongs to different intents. More precisely, the depth image $I$ and gripper pose $h$ are processed with a convolutional (Conv) encoder and a multilayer perceptron (MLP) encoder, respectively. Then the features are combined using concatenation operation and fed into three subsequent fully connected (FC) layers with 256, 256, and 128 neurons. The Conv encoder consists of one 1x1 convolutional layer followed by a global average pooling. The MLP encoder consists of one FC layer. We use the following cross-entropy loss to train the network, as shown in Eq. \ref{equ:loss}:
\begin{equation}
L=\frac{1}{N} \sum_{i} L_{i} = -\frac{1}{N} \sum_{i} \sum_{c=1}^{n} \{k_i = c\} \log \left(p_{i c}\right),
\label{equ:loss}
\end{equation}
where $c = (1,2,3,\ldots,n)$ is the class index of intents and $k_i$ is the intent class label of the given state. 

\subsection{Policy Learning with Intention Estimator}\label{sec:policy_rl}

After we train an intention estimator that can discern the intent of the given state, we distill an intrinsic reward from its prediction. The intrinsic reward allows the robot to follow the intent during policy learning in RL and is detailed below.

\textbf{Problem Formulation:} We formulate the picking problem as a Markov Decision Process (MDP). The MDP is defined by a state-space $S$, action space $A$, a function of reward $R(s_t, s_{t+1})$, and the transition probability $P(s_{t+1} |s_t , a_t)$. At time step $t$, a robot agent to pick objects observes the state $s_t$ and predicts an action $a_t$ based on current policy $\pi(a_t |s_t)$. The rewards from the environment and intention estimator are provided to the agent afterward and then transition to a new state $s_{t+1}$. RL aims to learn an optimal policy $\pi $ that selects actions that maximize its cumulative reward.

\textbf{Rewards with Intents:} 
The output probabilities from the intention estimator are used to design a reward function $r^{\prime}$ as follows:
\begin{equation}
r_{t}^{\prime} = p\{(k_i=t)|(s_{t+1})\},
\end{equation}
where $s_{t+1}$ is the state at time step $t+1$ and $p\{(k_i=t)|(s_{t+1})\}$ is a predicted probability between 0 and 1 representing the similarity between the current state and the $t^{th}$ intention $k_t$. \comment{The $k_i=t$ represents that we bundle time step $t$ with intent index $k_i$ to encourage the agent to follow the intents during training. Note the proposed method does not strictly limit agents to follow intents.} In order to grasp in finite steps, the episode length is fixed to the number of intents. 

Meanwhile, a task reward $r_{task}$ is given at the end of an episode, 10 for grasping one object successfully and 0 for otherwise. Thus the full reward function is defined as $r_{total} = r_{task} + r^{\prime}$. When the RL agent meets scenes that are never encountered in demonstrations, though the intrinsic reward $r^{\prime}$ from the intention estimator is almost zero, the agent can still explore on its own and obtain the task reward $r_{task}$ to learn the grasping policy in novel scenes.

\textbf{Policy Architecture:} The policy is trained with Proximal Policy Optimization (PPO) in simulation. PPO requires training a value network that forecasts the discounted sum of future rewards from the current state and a policy network that maps a current state to actions. The policy and value networks share the same state input, as shown in Fig. \ref{fig:fg2}. The state is defined as $s_t:= (I_t, h_t)$, where $I_t$ is a depth image with a resolution of 120×120 from the camera, and $h_t$ is the gripper pose at time step $t$ including the position, orientation, and closure status of the gripper. The policy and value networks share the same front-end network. $I_t$ is sequentially processed by three convolutional layers with kernel sizes of $8\times8$, $4\times4$, and $3\times3$, and $h_t$ is processed by one FC layer with eight neurons. Then, the concatenation of two extracted features is fed through the subsequent two FC layers with 64 neurons and split into two output layers: one for predicting the action and another for estimating the value. As a wrist-mounted camera on a real robot might capture things outside the workspace after acting $a_t$, to reduce the sim-to-real gap, we only update $h_t$ and always use the initial depth observation $I_0$ as the part of state $s_t$ instead of updating the depth observation over time.

\textbf{Actions:} In our environment, each policy action $a_t$ includes a gripper pose displacement and a vector to control the gripper closure. The gripper pose displacement is the difference between the initial pose and the desired one, encoded as $ (x_{t}^{\prime},y_{t}^{\prime},z_{t}^{\prime},\alpha_t^{\prime},\beta_t^{\prime},\gamma_t^{\prime})$, where $(x_t^{\prime},y_t^{\prime},z_t^{\prime})$ is the relative displacement and $(\alpha_t^{\prime},\beta_t^{\prime},\gamma_t^{\prime})$ are the rotations of the gripper about its $x$-, $y$- and $z$-axes. The one-hot vector to control the gripper closure is denoted as $\psi_t^{\prime}$. We discretize each action's coordinate according to the workspace. In addition, the episode will be terminated if $\psi_t$ is true. \comment{If terminated, the robot returns to its initial pose, receives new observations, and executes the next grasp, which provides a certain degree of ability for handling uncertainty and imperfect executions. For example, if objects slip from the hand during the last grasp trail, the robot can try again when it receives new observations after resetting to its initial pose.}

\subsection{Training Details}\label{sec:DC+AB_METHOD}

We train the policy in the Pybullet simulator\cite{coumans2021}. The training process consists of two stages. First, we learn the intention estimator from grasp demos, and then the RL agent explores and learns the grasping policy with the aid of the intrinsic reward constructed by the intention estimator. To train the intention estimator, we generate 10000 grasps in the simulation for each provided grasp type using the method described in Sec. \ref{sec:learn_intention}. Each encoded grasps have four poses. We use $n=3$ as the number of intents. Intuitively, when the gripper closes, it often represents a shift of intention, and thus we use $\mu = 5$ to increase the influence of such activities in the dissimilarity calculation. A total of 30000 grasps were used to train the intention network with cross-entropy loss. The Adam optimizer was employed, starting with a learning rate of 0.001. One hundred epochs are performed, and the learning rate is halved every ten epochs during training. 

\begin{figure}[!t]
\vspace{0.15cm}
    \centering
    \begin{subfigure}{\linewidth}
        \begin{overpic}[width=\linewidth]{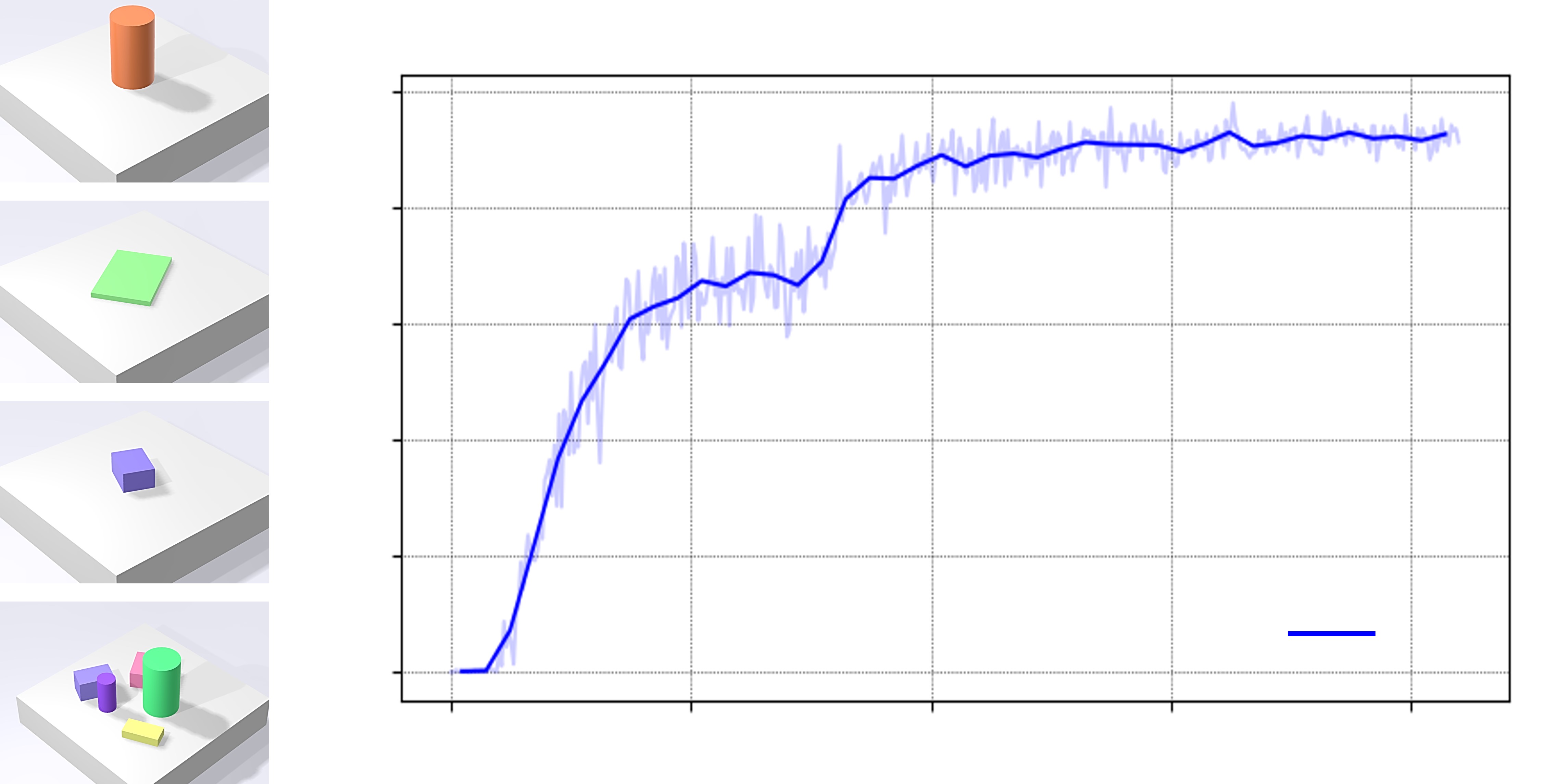}

        \put(28.7,2.25) {\scriptsize 0}
        \put(44,2.25) {\scriptsize 5}
        \put(59,2.25) {\scriptsize 10}
        \put(74.3,2.25) {\scriptsize 15}
        \put(89.7,2.25) {\scriptsize 20}

        \put(17,20){ {\rotatebox{90}{\scriptsize Success rate (\%)}}}
        \put(22.7, 6.5) {\scriptsize 0}
        \put(22,14) {\scriptsize 20}
        \put(22,21.5) {\scriptsize 40}
        \put(22,29) {\scriptsize 60}
        \put(22,36.5) {\scriptsize 80}
        \put(21,44) {\scriptsize 100}
        
        \put(52,-0.5) {\scriptsize Episodes (Millions)}
        \put(8,-4.5) {\footnotesize (a)}
        \put(59,-4.5) {\footnotesize (b)}
        \put(90,9) {\scriptsize PPO}
        \put(48,-44) {\footnotesize (c)}
        \end{overpic}
    \end{subfigure}
    
    \vspace{-0.4cm}
    \begin{subfigure}{\linewidth}
        \resizebox{\linewidth}{!}{
            \begin{tabular}{lcccc} 
            \hline
            \multicolumn{1}{c}{\multirow{2}{*}{Method}} & \multicolumn{3}{c}{Success Rate} \\ 
            \cline{2-4}
            \multicolumn{1}{c}{} & \begin{tabular}[c]{@{}c@{}}Single Object\\(Similar Scene)\end{tabular} & \begin{tabular}[c]{@{}c@{}}Clutter\\(Unseen Scene)\end{tabular} & Total \\ 
            \hline
            BC & 83.5\% & 23.1\% & 53.3\% \\
            BC-perfect demo & 88.7\% & 26.8\% & 57.8\% \\
            \textbf{Ours} & \textbf{98.3\%} & \textbf{91.2\%} &\textbf{ 94.8\%} \\
            Ours-perfect demo & 98.7\% & 91.5\% & 95.1\% \\
            Ours-w/o intent & 61.2\% & 86.6\% & 73.9\% \\
            \hline
            \end{tabular}
        }
    \end{subfigure}
    \vspace{0.2cm} 
    \caption{(a): Examples of environments. The first three rows are similar environments to the demo, the clutter of the last row is not in the demo (b): Success rate curve of our policy training. (c): Simulation results with different element choices of our method (Ours) and behavior cloning (BC). }
    \label{fig:fg4}
\vspace{-0.5cm}
\end{figure}
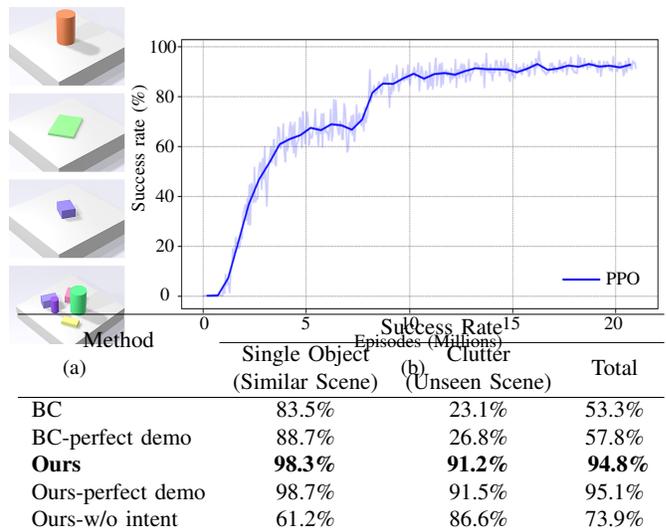

During the RL policy learning phase, a pool of 64 robots generates training episodes by downloading the current policy parameters every 10 epochs from the optimizer. In each environment, random objects were placed in the workspace with random poses. Only cuboids, cylinders, and their variants with different aspect ratios are used during training, as shown in Fig. \ref{fig:fg4}(a). The robot then collects the episodes in the simulation, during which the simulator automatically determines the task reward, and the estimator provides the intrinsic reward. If the workspace is empty or an object is dropped, the environment will be reset, at which point objects with random poses will fall into the workspace again. Finally, the resulting episodes are sent back to the optimizer. The Adam optimizer with a learning rate of $10^{-4}$ is used. We also deploy domain randomization to make the learned policy robust to a range of real-world conditions. Fig. \ref{fig:fg4}(b) shows the learning curve for the final model training.

\subsection{Simulation Results}\label{sec: simulation results}

After training, the policy network alone is deployed to the robot in both simulation and the real world. In simulation experiments, we set up the following environments: a) Scenes in demonstrations (Similar Scene): environment constructed with a single object but in new configurations, including object friction and mass. b) Scenes not in demonstrations (Unseen Scene): a cluttered scene with multiple objects, \comment{where grasping policies can be found by self-exploration.} Fig. \ref{fig:fg4}(c) summarizes the result tested on similar and unseen scenes in simulation. The learned policy from the final model (denoted as $Ours$) is able to grasp the object with success rates of 98.3\% in the similar scene and 91.2\% in the unseen scene (row 3 in Fig. \ref{fig:fg4}(c)). In contrast, removing the intent estimator from RL policy learning (denoted as $Ours$-$w/o$ $intent$), the success rates are considerably lower (row 5 vs. row 3 in Fig. \ref{fig:fg4}(c) because pure RL exploration cannot find a successful grasping policy for thin objects such as cards. Meanwhile, the policy directly learned by behavior cloning (denoted as $BC$) using the same demonstrations performs better than $Ours$-$w/o$ $intent$ but lower than $Ours$. This validates our hypothesis that learning from intent (row3 in Fig. \ref{fig:fg4}(c)) can help the agent learn complex and better policies while retaining its ability to explore unseen scenarios beyond directly cloning policies (row 1 in Fig. \ref{fig:fg4}(c)) or exploring entirely on its own (row 5 in Fig. \ref{fig:fg4}(c)). Moreover, the method of behavior cloning achieves poor performance in the unseen scene. In comparison, our model generalizes well to the novel scene. 

We also investigate the impact of demonstration quality on policy learning. A total of 12.3\% of demos for training intent estimators fail. We remove these imperfect demonstrations and use the remaining perfect demos to learn the policy using behavior cloning and our method (row 2 and row 4 in Fig. \ref{fig:fg4}(c)). We observe that by using perfect demonstrations, the success rate of behavior cloning increases by more than 5\% compared to using imperfect demonstrations (row 2 vs. row 1 in Fig. \ref{fig:fg4}(c)). In contrast, our method does not rely on the quality of the demonstration. It achieves comparable performance with the one learned with perfect demonstrations (row 3 vs. row 4 in Fig. \ref{fig:fg4}(c)). Because when the intent estimator's guidance is biased due to imperfect demonstrations, the RL agent can revise the policy through self-exploration, illustrating the superiority of learning from intents rather than direct cloning atomic actions. \comment{Such ability also helps agents to learn in unseen scenes that the agent seamlessly switches to self-exploration when meeting novel scenes, allowing agents to obtain policies in scenes that are not demonstrated.} Real-world experiment results are presented in Sec. \ref{sec: real_exp_total}.

\section{Real-World Experiments}\label{sec: real_exp_total}

We executed a set of experiments to evaluate our system in the real world. The code of the presented work is available \href{https://robotll.github.io/LearnfromIntents/}{https://robotll.github.io/LearnfromIntents/}

\subsection{Hardware Setting}

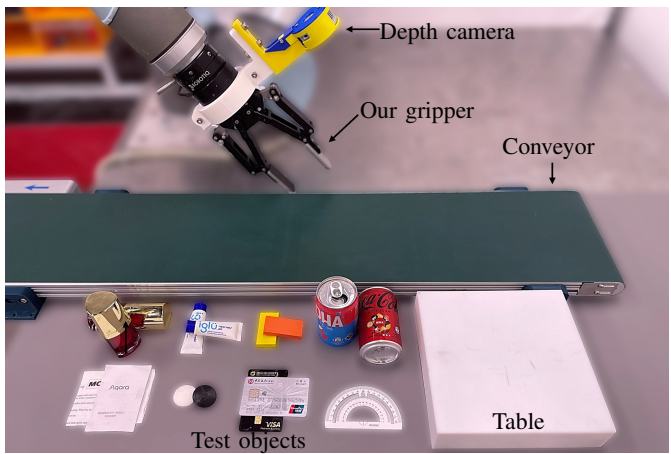
\begin{figure}[!ht]
\vspace{-0.25cm}
    \centering
    \begin{overpic}[width=\linewidth]{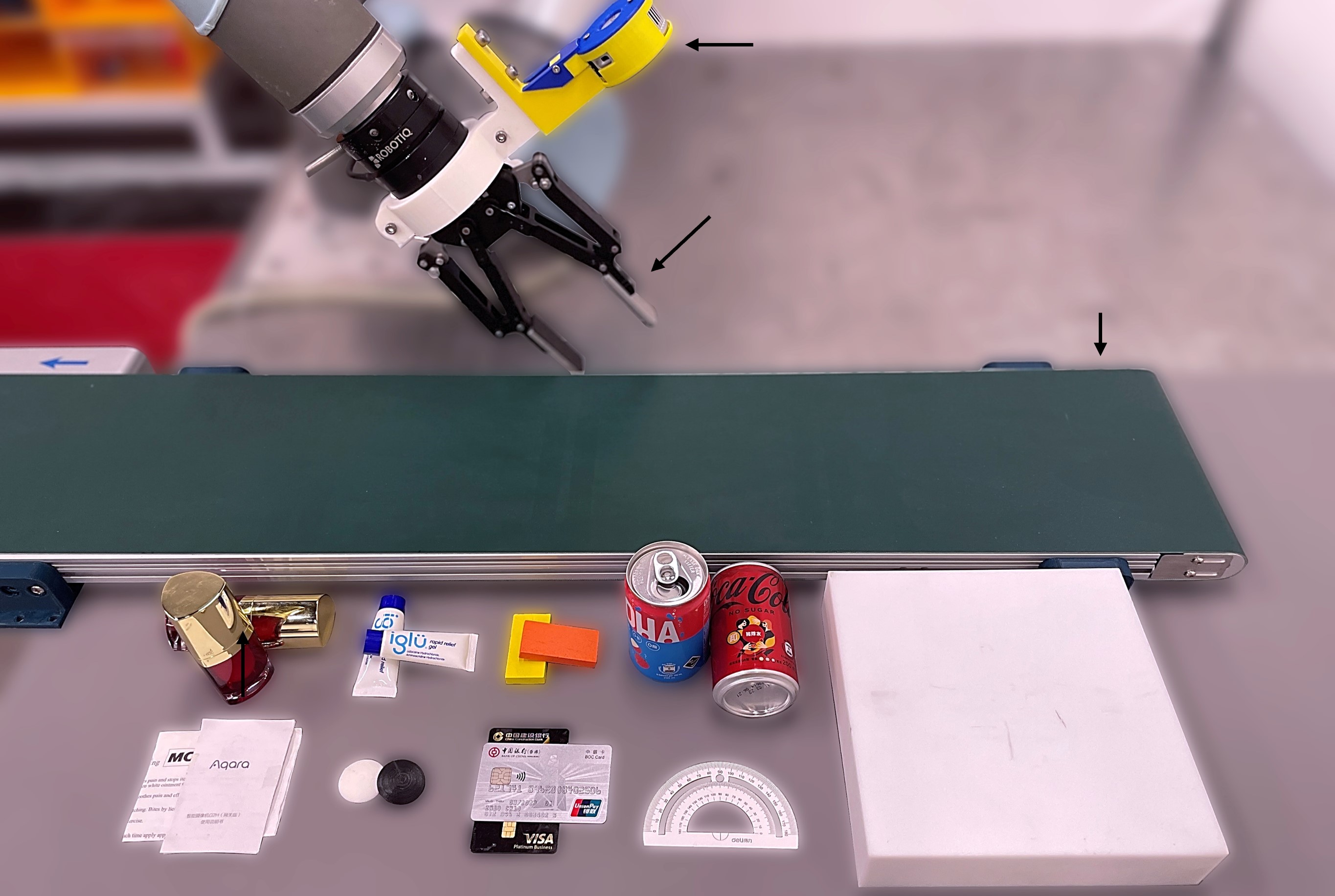}
    \put(56.5,62.5) {\small {Depth camera}}
    \put(53.5,50.5) {\small {Our gripper}}
    \put(73.5,3.5) {\small {Table}}
    \put(75,45) {\small {Conveyor}}
    \put(28,0.7) {\small {Test objects}}
    \end{overpic}
    \caption{Our hardware setting for real-world experiments.}
    \label{fig:fg5}
\vspace{-0.25cm}
\end{figure}

As shown in Fig. \ref{fig:fg5}, we deployed the learned policy on an off-the-shelf robotic grasping platform, including a 6-DOF robot arm equipped with a robotiq140 parallel gripper and an Intel L515 depth camera.

\begin{table*}[!t]
\vspace{0.25cm}
\centering
\caption{Results of experiments in the real world.}
\label{tab:real}

\resizebox{\linewidth}{!}{
\begin{threeparttable}
\begin{tabular}{cccccccccccccccccccc}
\hline

\multirow{6}{*}{ENV} & \multirow{6}{*}{Method} & \multicolumn{2}{c}{Credit Card} & \multicolumn{2}{c}{User manual} & \multicolumn{2}{c}{Protractor} & \multicolumn{2}{c}{Domino} & \multicolumn{2}{c}{Tube} & \multicolumn{2}{c}{Go Stone} & \multicolumn{2}{c}{Soda Can} & \multicolumn{2}{c}{Cosmetic Jars} & \multicolumn{2}{c}{Dense Clutter} \\ 

 &  & \multicolumn{2}{c}{\multirow{4}{*}{\includegraphics[scale=0.18]{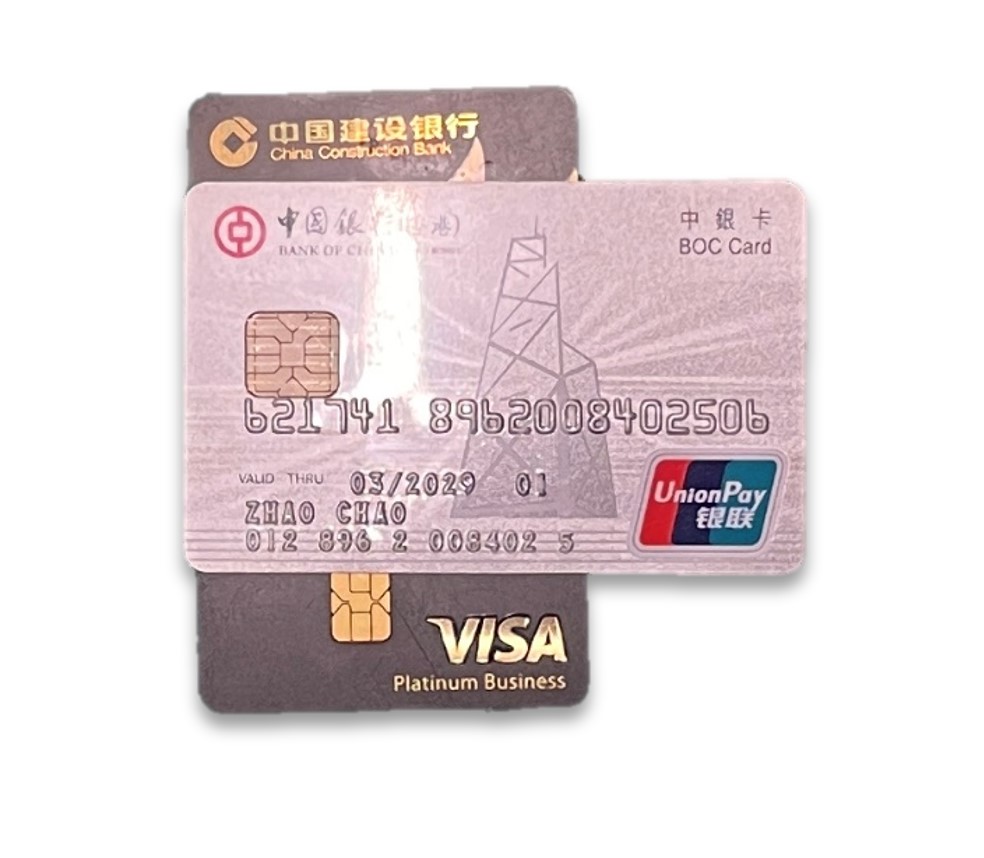}}} & \multicolumn{2}{c}{\multirow{4}{*}{\includegraphics[scale=0.18]{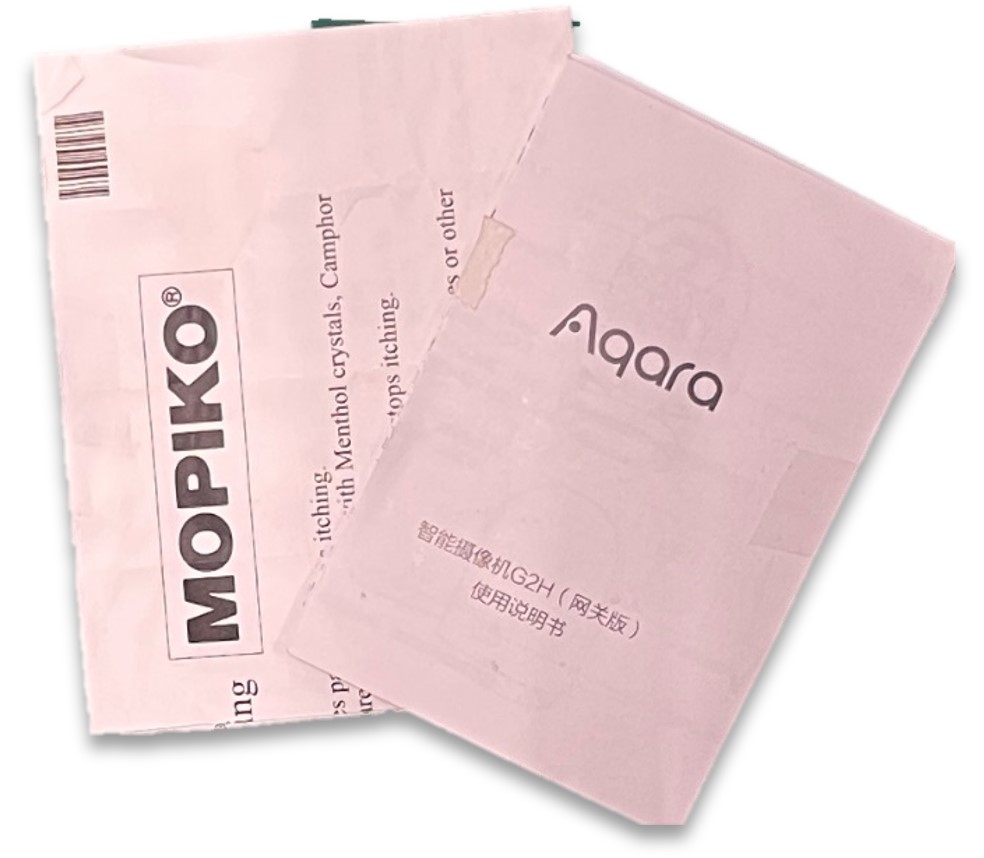}}} & \multicolumn{2}{c}{\multirow{4}{*}{\includegraphics[scale=0.18]{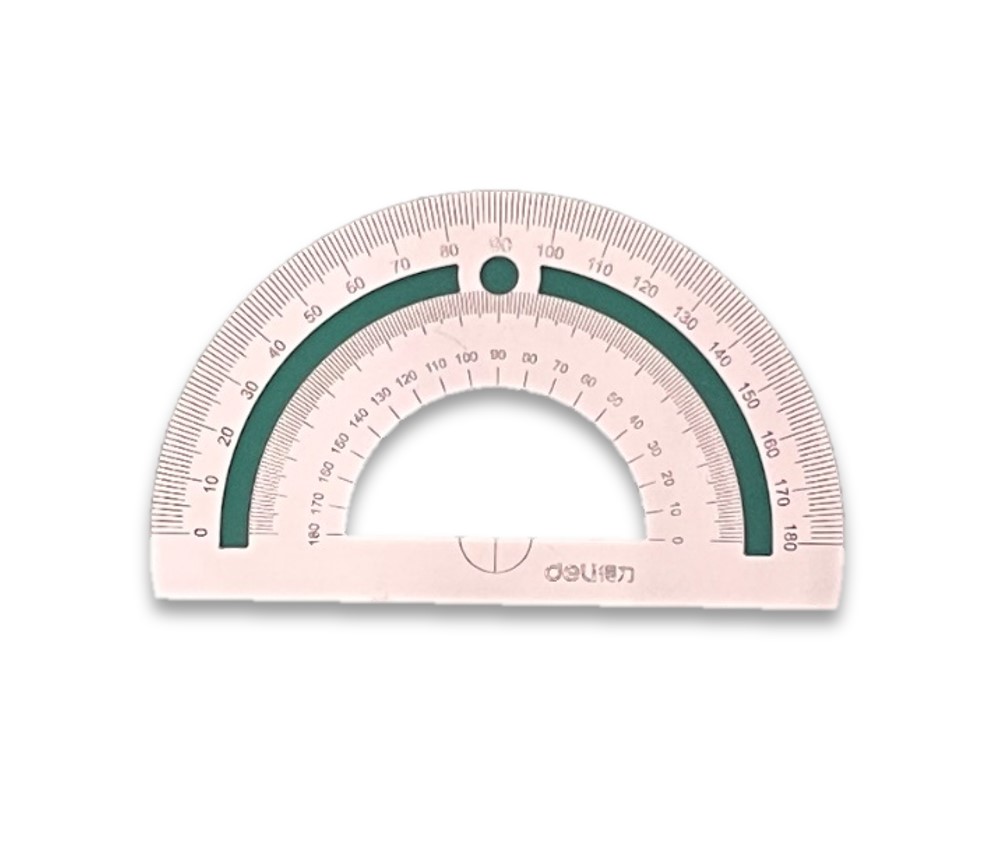}}} & \multicolumn{2}{c}{\multirow{4}{*}{\includegraphics[scale=0.18]{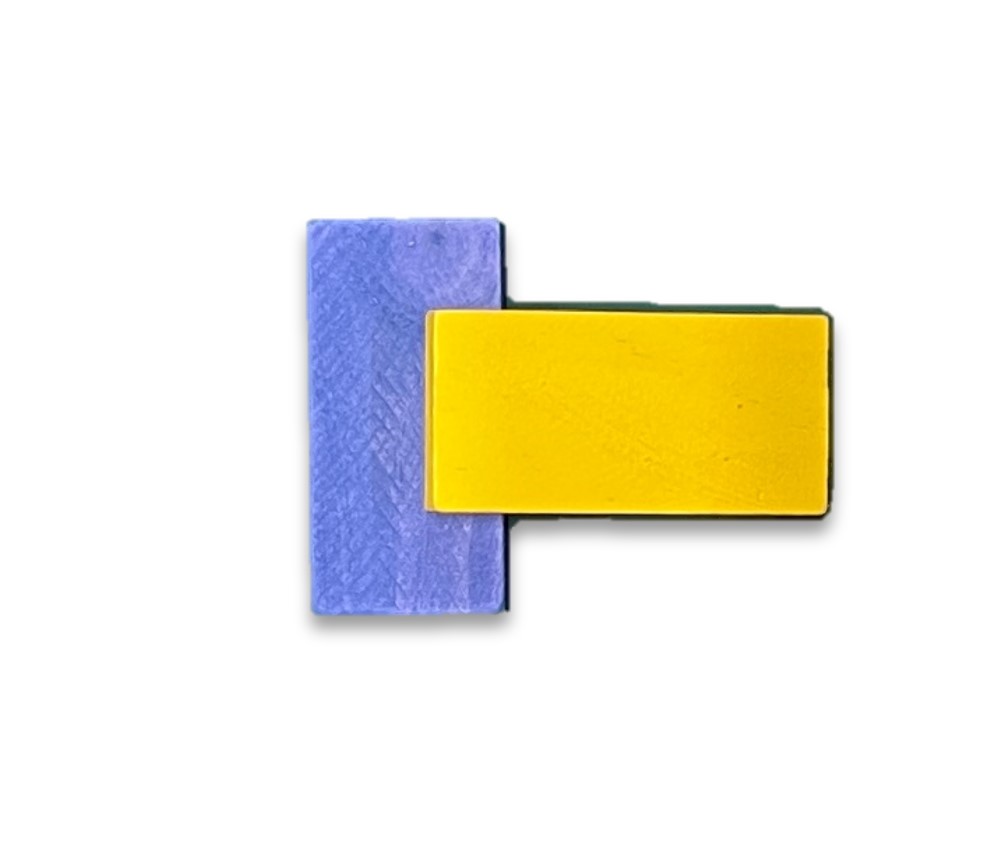}}} & \multicolumn{2}{c}{\multirow{4}{*}{\includegraphics[scale=0.18]{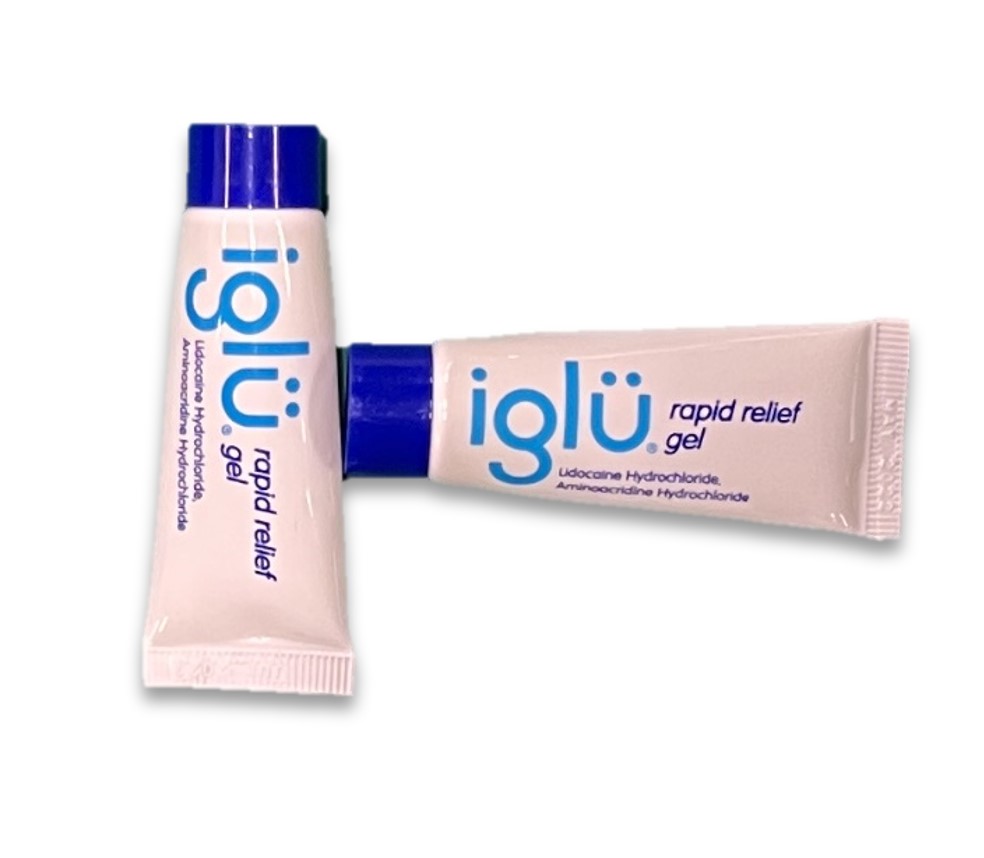}}} & \multicolumn{2}{c}{\multirow{4}{*}{\includegraphics[scale=0.18]{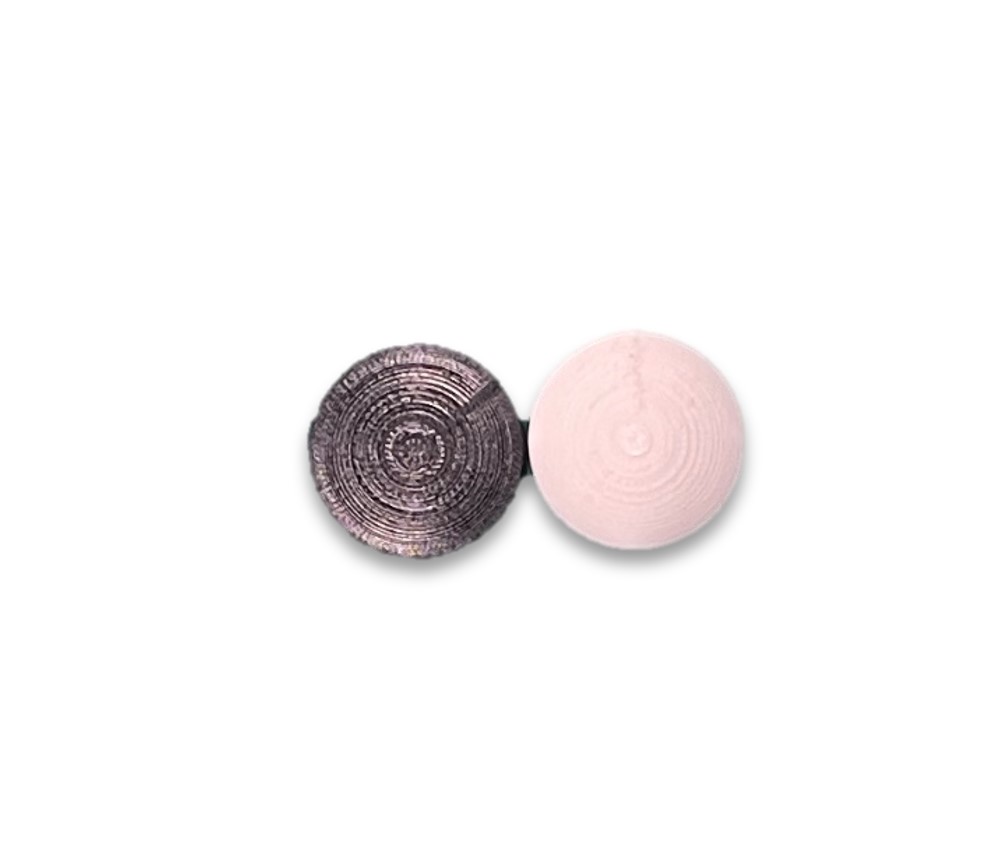}}} & \multicolumn{2}{c}{\multirow{4}{*}{\includegraphics[scale=0.18]{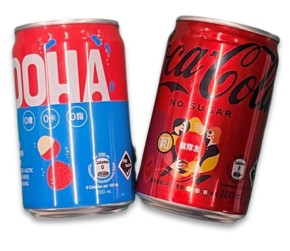}}} & \multicolumn{2}{c}{\multirow{4}{*}{\includegraphics[scale=0.18]{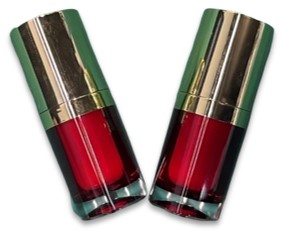}}} & \multicolumn{2}{c}{\multirow{4}{*}{\includegraphics[scale=0.18]{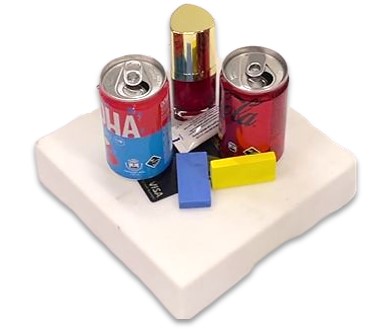}}} \\[.25ex]
 &  & \multicolumn{2}{c}{} & \multicolumn{2}{c}{} & \multicolumn{2}{c}{} & \multicolumn{2}{c}{} & \multicolumn{2}{c}{} & \multicolumn{2}{c}{} & \multicolumn{2}{c}{} & \multicolumn{2}{c}{} & \multicolumn{2}{c}{} \\[1ex]
 &  & \multicolumn{2}{c}{} & \multicolumn{2}{c}{} & \multicolumn{2}{c}{} & \multicolumn{2}{c}{} & \multicolumn{2}{c}{} & \multicolumn{2}{c}{} & \multicolumn{2}{c}{} & \multicolumn{2}{c}{} & \multicolumn{2}{c}{} \\[.25ex]
 &  & SR & PPH & SR & PPH & SR & PPH & SR & PPH & SR & PPH & SR & PPH & SR & PPH & SR & PPH & SR & PPH \\[.25ex] \hline

 &  &  &  &  &  &  &  &  &  &  &  &  &  &  &  &  &  &  &  \\[-1.5ex]
Table & VPN \cite{cai2022real} & 0\% & - & 0\% & - & 0\% & - &\textbf{ 98\%} & 101 & \textbf{98\%} & 101 & 22\% & 23 & 80\% & 82 & 84\% & 87 & 38\% & 39 \\[1ex]
\multirow{3}{*}{\includegraphics[scale=0.15]{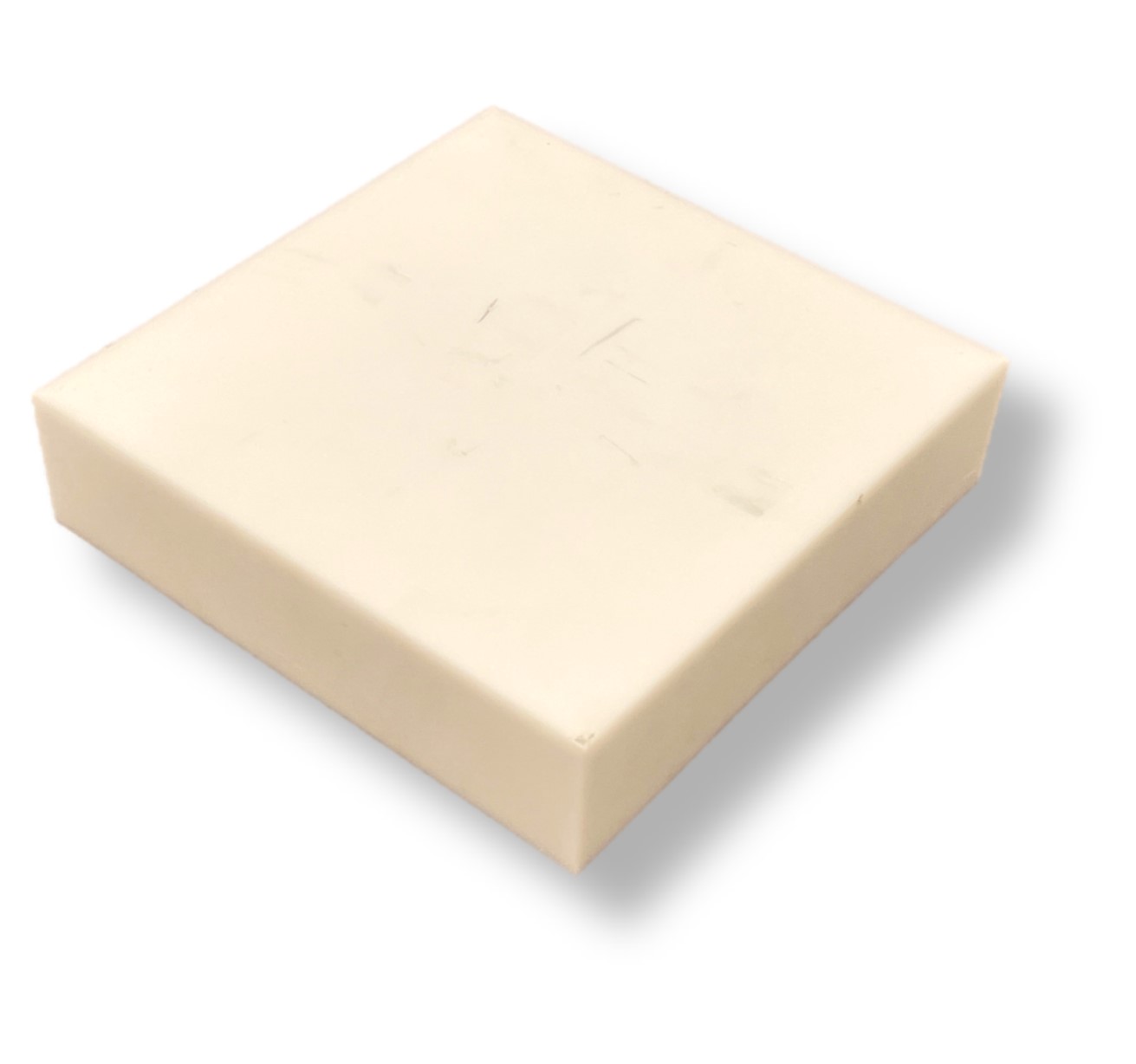}} 
 & Planar \cite{zeng2018robotic} & 0\% & - & 0\% & - & 0\% & - & 96\% & 346 & 92\% & 331 & 64\% & 230 & 90\% & 324 & 82\% & 295 & 66\% & 238 \\[1ex]
 & BC & 50\% & 178 & 46\% & 163 & 18\% & 64 & 90\% & 320 & 84\% & 298 & 52\% & 185 & 96\% & 341 & 86\% & 305 & 12\% & 43 \\[1ex] 
 & \textbf{Ours} & \textbf{82\%} & \textbf{291} &\textbf{ 76\%} &\textbf{270} & \textbf{64\%} & \textbf{227} &\textbf{ 98\%} & \textbf{348} & 94\% & \textbf{334}  &\textbf{ 68\%} &\textbf{ 241} & \textbf{96\%} & \textbf{341} & \textbf{92\%} & \textbf{327} & \textbf{82\%} & \textbf{291 }\\[1ex] \hline

 &  &  &  &  &  &  &  &  &  &  &  &  &  &  &  &  &  &  &  \\[-1.5ex]
\multirow{3}{*}{\includegraphics[scale=0.15]{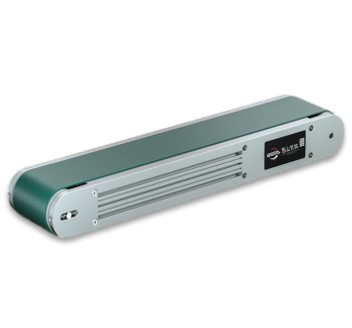}} 
 & VPN \cite{cai2022real} & 0\% & - & 0\% & - & 0\% & - & \textbf{98\%} & 101 & \textbf{98\%} & 101 & 22\% & 23 & 84\% & 87 & 82\% & 84 & 52\% & 54 \\[1ex]
 & BC & 38\% & 135 & 34\% & 121 & 12\% & 43 & 90\% & 320 & 84\% & 298 & 40\% & 142 & 92\% & 327 & 86\% & 305 & 10\% & 36 \\[1ex]
 & \textbf{Ours }& \textbf{80\%} & \textbf{284} & \textbf{68\%} & \textbf{241} & \textbf{60\%} & \textbf{213} & \textbf{98\%} & \textbf{348} & 92\% &\textbf{ 327} &\textbf{ 64\%} & \textbf{227 }&\textbf{ 98\%} & \textbf{348} & \textbf{90\%} &\textbf{ 320} & \textbf{78\%} & \textbf{277 }\\[1ex] \hline

\end{tabular}
\begin{tablenotes}
\item[*] SR stands for Success rate. $^{**}$ Dense Clutter: Mixed objects on the cluttered table or the conveyor belt.
\end{tablenotes}
\end{threeparttable}
}
\vspace{-0.2cm}
\end{table*}

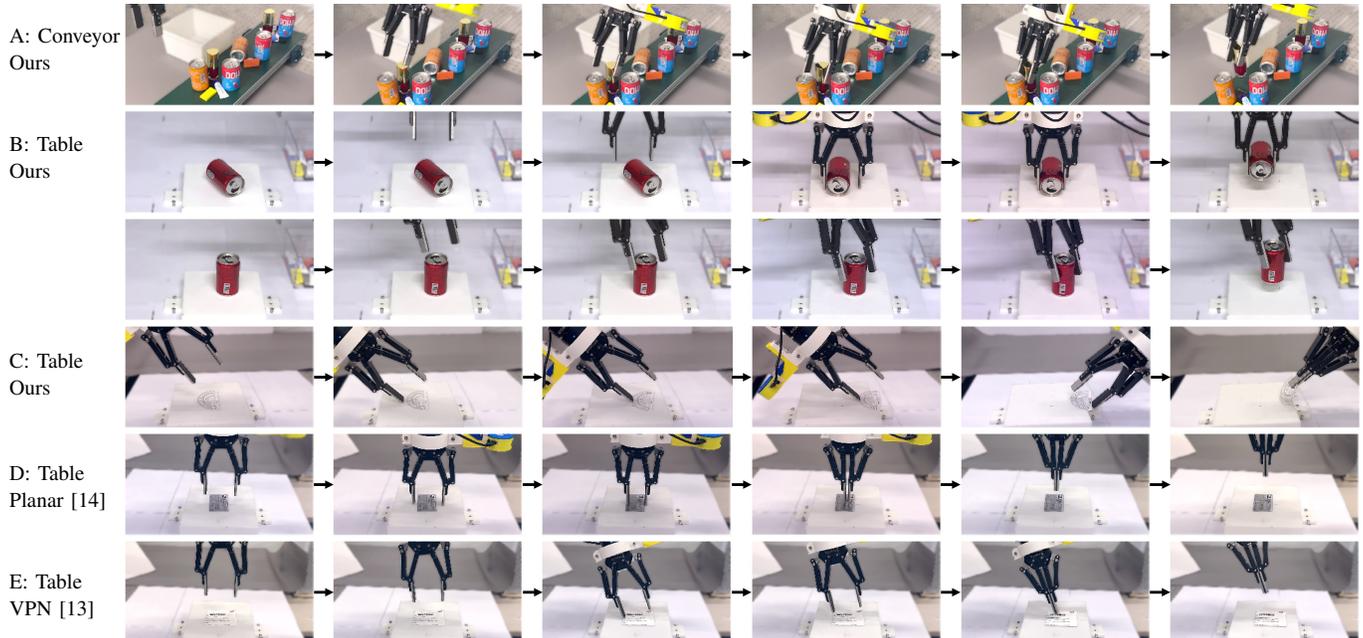
\begin{figure*}[!htbp]
    \centering
    \begin{overpic}[width=\linewidth]{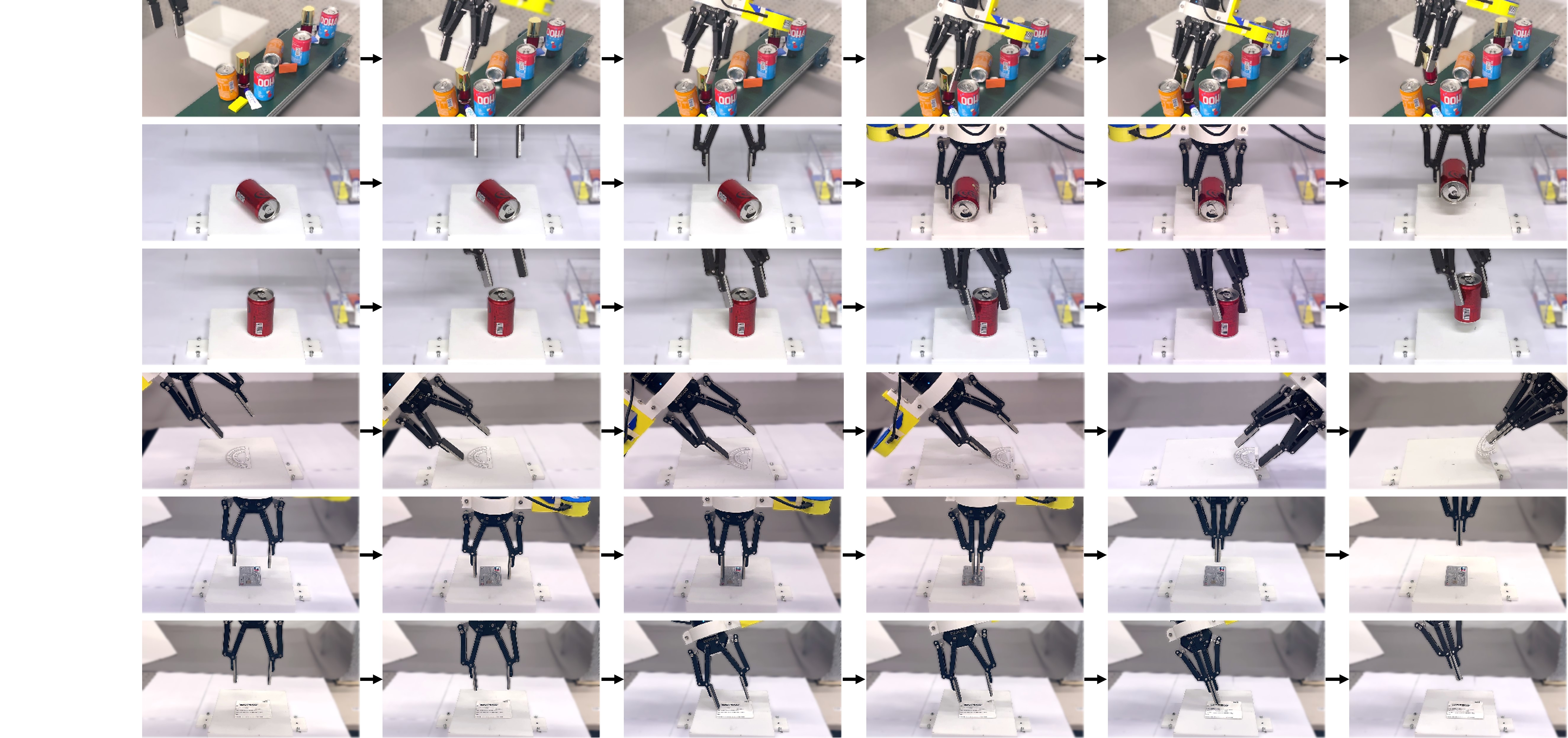}
    
    \put(2,55) {\scriptsize Conveyor}
    
    \put(0.5,44.3) {\footnotesize A: Conveyor}
    \put(0.5,42.3) {\footnotesize Ours}
    
    \put(0.5,36.3) {\footnotesize B: Table}
    \put(0.5,34.3) {\footnotesize Ours}
    
    \put(0.5,20.3) {\footnotesize C: Table}
    \put(0.5,18.3) {\footnotesize Ours}
    
    \put(0.5,12) {\footnotesize D: Table}
    \put(0.5,10) {\footnotesize  Planar \cite{zeng2018robotic}}
    
    \put(0.5,4) {\footnotesize E: Table}
    \put(0.5,2) {\footnotesize  VPN \cite{cai2022real}} 

    \end{overpic}
    \caption{
    A: our method (Ours) grasps from a cluttered conveyor; Successful grasp. 
    B: our method (Ours) responds to a soda can with different poses; Successful grasps with robust grasping behavior. 
    C: our method (Ours) grasps a protractor; Successfully grasp. 
    D: top-down grasp (Planar \cite{zeng2018robotic}) cannot pick the credit card. The card slips out of the fingertip.
    E: 6-DoF grasp synthesis method (VPN \cite{cai2022real}) fails to grasp the user manual due to the collision.
    }
    
    \label{fig:fg6}
\vspace{-0.5cm}
\end{figure*}

\subsection{Real Robot Experiment}\label{sec:real_robot_exp}

In this section, we quantitatively evaluated our picking system and other state-of-the-art methods with two protocols. In the first protocol, which we refer to as “isolated object grasping”, the robot attempted to grasp a single object lying in the workspace. We also used a second protocol where the robot cleaned a pile of mixed objects randomly dumped into the workspace. This test was more challenging as the robot had to avoid collisions with other objects while grasping. We used two metrics for evaluation: successful picks per attempt (Success Rate) and picks per hour (PPH). A successful grasp is grasping only one object and not pushing any other objects out of the workspace. Tab. \ref{tab:real} summarizes the results of the learned policy in the real world. 

We first examined the performance of our policy with the first protocol (col. 1-8 in Tab. \ref{tab:real}) on the table environment. Our method (Ours) obtained success rates of over 90\% for dominos, tubes, cans, and cosmetics. For the most challenging objects, including cards, user manuals, protractors, and Go stones, our method achieved success rates of over 80\% for cards and over 60\% for the other objects. In contrast, the state-of-the-art 6-DoF grasp synthesis method (VPN) \cite{cai2022real} and the learning-based planar grasping method (Planar) \cite{zeng2018robotic} could not successfully grasp cards, user manuals, and protractors. Notably, when testing dominos, tubes, and Go stones using the VPN baseline, we manually select top-down grasp poses; otherwise, the VPN method cannot detect a feasible grasp for these objects. Meanwhile, the behavior cloning (BC) method performed worse with all test objects. 

We then evaluated our learned policy on the cluttered table populated by multiple objects (column 9 in Tab. \ref{tab:real}). Our method stably obtained a success rate of 82\% in the challenging dense clutter. This level of performance is beyond other baselines. Also, the success rates of behavior cloning dropped below 15\% on mixed objects due to the inherent compounding error and distribution shift. From Tab. \ref{tab:real}, the protractor and the Go stone are the most challenging to grasp among test objects. We hypothesize that this happened because protractors and Go stone have complex geometries and dynamics different from the training objects, increasing the difficulty of generalization. The other methods also perform less effectively. 

At last, to emphasize the generalization ability of our learned policy and the value of using an off-the-shelf parallel gripper alone to grasp objects in broad categories, we also focused on comparing the presented approach with other methods in a conveyor environment common in industry (see Fig. \ref{fig:fg6}A). The belt on the conveyor has higher friction than the table environment and is elastic. In addition, the significant variation of material properties over the surface adds extra noise to the depth camera. Overall, our method reported in the third row still achieved a higher grasp success rate (except for the tube) and PPH in all conditions. The tube has a lower success rate as it has a different non-centrosymmetric shape, making it easier to grasp from the head rather than the object's center. In contrast, all training objects are centrosymmetric and do not have such a characteristic.

The learned policy manifests a dexterous behavior, as shown in Fig. \ref{fig:fg6}. The robot approaches the protractor and continues interacting to reach a state that is feasible to grasp (see Fig. \ref{fig:fg6}C). This distinguishes the presented approach from other exploration-based methods, which confine the policy to a format of approaching the object with a certain pose, closing the finger, and avoiding interaction with objects (see Fig. \ref{fig:fg6}D and \ref{fig:fg6}E). We can also observe that the behavior cloning method performs poorly due to the distribution shift issue, further showing the significance of learning from intentions. Note also that the policy learned by our method is more robust and not tied to particular objects. Fig. \ref{fig:fg6}B shows the learned policy responding to different poses of the same object. The policy identifies purely from observations and adopts different strategies. Such behavior is not specified during training in any way and is discovered by itself. Our training environment features only simple rigid objects, with no complex geometry or compliance, such as  protractors and user manuals. Nevertheless, the learned policy successfully meets the diversity of real-world conditions encountered at deployment.

\subsection{Further analysis of generalization}\label{sec:revise}

\begin{table}
\centering
\caption{ {Analysis of generalization}}
\label{tab: extra}
\resizebox{\linewidth}{!}{
\begin{threeparttable}
\begin{tabular}{ccccccccc} 
\hline
\multirow{2}{*}{ {Method}} & \multicolumn{2}{c}{ {Use extra objects?}} & \multicolumn{2}{c}{ {Protractor}} & \multicolumn{2}{c}{ {Go stone}} & \multicolumn{2}{c}{ {Tube}} \\
 &  {Phase A\textsuperscript{*}} &  {Phase B\textsuperscript{**}} &  {Sim} &  {Real} &  {Sim} &  {Real} &  {Sim} &  {Real} \\ 
\hline
 {Ours} &  {No} &  {No} &  {67\%} &  {64\%} &  {72\%} &  {68\%} &  {95\%} &  {94\%} \\
 {Ours-extra} &  {No} &  {Yes} &  {97\%} &  {84\%} &  {98\%} &  {88\%} &  {98\%} &  {98\%} \\
 {Ours-w/o intent} &  {$~~~-^{***}$} &  {Yes} &  {0\%} &  {0\%} &  {93\%} &  {84\%} &  {98\%} &  {98\%} \\
\hline
\end{tabular}
\begin{tablenotes}
\item[\scriptsize{*}] \scriptsize{Intention estimator learning stage. $^{**}$ Policy learning stage. $^{***}$ Phase A excluded.}
\end{tablenotes}
\end{threeparttable}
}
\vspace{-0.65cm}
\end{table}

In this section, we investigate 1) the effect on the success rate of adding novel objects, which perform relatively poorly during real-world testing, into the policy training phase; 2) how well the intention estimator, trained on only cube and cylinder, generalizes to different objects. For the first question, we add models of the protractor, Go stone, and tube to the policy learning phase (i.e., Phase B in Fig. \ref{fig:fg2}) and train with our proposed method (denoted as \textit{Ours-extra}). Qualitative real-world and simulation results (row 2 vs. row 1) show that the success rates of these objects can be improved by adding their models to training. For the second question, we learn the policy with extra objects and without using the intention estimator, denoted as Ours-w/o intent in Tab. \ref{tab: extra}, and compare its performance with \textit{Ours-extra}. The results (row 3 vs. row 2) show that the intention estimator successfully generalizes to objects that differ from those used in the intention estimator training phase. \textit{Ours-extra} achieves an over 90\% success rate for grasping the protractor, while \textit{Ours-w/o intent}, which only purely relies on self-exploration, cannot grasp the protractor successfully. Meanwhile, \textit{Ours-extra} achieves higher performance for the Go stone than \textit{Ours-w/o intent}. Both results show the successful generalization of the intention estimator to different novel objects. The results also show that without the guidance of the intention estimator, RL agents' self-exploration cannot discover a successful policy to grasp challenging thin objects (e.g., protractor).

\section{Discussion and Future Work}

Unlike other state-of-the-art methods, our approach mimics the human learning process, which abstracts and learns intent from demonstrated grasps, and then develops grasping policies through self-exploration. Despite the challenging objects, our method achieves up to 82\% success in the dense clutter. While a set of demoed grasps need to be collected, all grasps are collected automatically in the simulation based on three human-encoded grasps.
This minimizes the workload on humans. On the other hand, our approach leverages the intent as a reward during RL policy training without imitating detailed motion. Hence, it can learn to react to environments and scene settings not included in the demos. 

We see several limitations and opportunities for future research. First, our result describes a far wider range of objects, which achieves substantial improvements over other approaches. 
Future research could extend the present work to include grasping deformable objects. Another hint is that we hypothesize that diverse environments and demos could extend the presented work to long-horizon tasks since the proposed methodology is generic concerning the tasks. Finally, the presented work relies on human-encoded grasps to learn complex policies that self-exploration cannot discover. This is a significant advantage in that some grasping policies are hard to discover with pure exploration.
Nevertheless, humans can easily learn behavior from videos or descriptions in books instead of human-encoded movement. A major opportunity for future studies will be to extend the proposed work to develop a method that can directly learn grasping policies from video or language descriptions.

\normalem
\bibliographystyle{ieeetr}
\bibliography{references}

\end{document}